\documentclass[letterpaper, 10 pt, conference]{ieeeconf}  

\IEEEoverridecommandlockouts                              
\usepackage[T1]{fontenc}

\usepackage[noadjust]{cite}

\usepackage{amsmath}
\usepackage{yhmath}
\usepackage{mathtools} 
\usepackage{amssymb} 
\usepackage{sidecap}
\usepackage{graphicx}
\usepackage{subcaption}
\usepackage{multirow}
\usepackage{epsfig}
\usepackage{psfrag}
\usepackage{caption}
\usepackage[super]{nth} 
\usepackage[normalem]{ulem} 
\usepackage{color}
\usepackage{xcolor} 
\usepackage{booktabs} 
\usepackage{siunitx} 
\usepackage{booktabs} 
\usepackage{colortbl} 
\usepackage{textcomp} 
\usepackage{tikz}
\usepackage[noadjust]{cite} 

\usepackage{wasysym}  
\usepackage{fancyhdr} 
\usepackage{color}
\usepackage{hyperref}  
\definecolor{darkblue}{rgb}{0.0,0.0,1}
\hypersetup{colorlinks=false,breaklinks,
            linkcolor=darkblue,urlcolor=darkblue,
            anchorcolor=darkblue,citecolor=darkblue}

\newcommand{\RNum}[1]{\uppercase\expandafter{\romannumeral #1\relax}} 

\newcommand{\realfield}[1]{\hbox{I \kern -.5em R}^{#1}}
\newcommand {\mb}[1]{\mathbf{#1}}
\newcommand {\bs}[1]{\boldsymbol{#1}}

\newcommand{\Rot}[2]{{^{#1}\mathbf{R}}_{#2}}  
\newcommand{\T}{^{\mathrm{T}}}  

\definecolor{LightGray}{gray}{0.9}
\newcolumntype{a}{>{\columncolor{LightGray}}l}
\newcommand*\circled[1]{\tikz[baseline=(char.base)]{\node[circle,minimum size=7pt,draw=black,inner sep=0.5pt](char){\scriptsize #1};}}

\newcommand{\PreserveBackslash}[1]{\let\temp=\\#1\let\\=\temp}
\newcolumntype{C}[1]{>{\PreserveBackslash\centering}p{#1}}
\newcolumntype{L}[1]{>{\PreserveBackslash}l{#1}} 
\usepackage{color}  

\usepackage[textsize=scriptsize, bordercolor=white,backgroundcolor=gray!30,linecolor=black,colorinlistoftodos]{todonotes}  
\usepackage[normalem]{ulem} 
\usepackage{soul} 

\title{\LARGE \bf
Task and Configuration Space Compliance of\\ Continuum Robots via Lie Group and Modal Shape Formulations}

\author{~Andrew~L.~Orekhov$^{1,2}$,~Garrison L.~H. Johnston$^{1}$,~Nabil~Simaan$^{1}$$^{\dag}$ 
\thanks{$\dag$ Corresponding author}
\thanks{$^{1}$Department of Mechanical Engineering, Vanderbilt University, Nashville, TN. email: {\tt aorekhov@andrew.cmu.edu, (garrison.l.johnston, nabil.simaan) @vanderbilt.edu}}
\thanks{$^{2}$The Robotics Institute, Carnegie Mellon University, Pittsburgh, PA.}
\thanks{This work was supported by NSF award \#1734461 and by Vanderbilt University internal funds. A. Orekhov was partially supported by the NSF Graduate Research Fellowship under \#DGE-1445197. }
}

\graphicspath{{fig}}
\usepackage{balance}
\begin{document}

\maketitle
\thispagestyle{empty}

\thispagestyle{fancy}
\fancyhf{}
\renewcommand{\headrulewidth}{0pt}
\lhead{2023 IEEE/RSJ International Conference on Intelligent Robots and Systems (IROS). Accepted Version. }
\rfoot{\centering \scriptsize \copyright 2023 IEEE. Personal use of this material is permitted. Permission from IEEE must be obtained for all other uses, in any current or future media, including reprinting/republishing this material for advertising or promotional purposes, creating new collective works, for resale or redistribution to servers or lists, or reuse of any copyrighted component of this work in other works.}

\pagestyle{empty}
\begin{abstract}
Continuum robots suffer large deflections due to internal and external forces. Accurate modeling of their passive compliance is necessary for accurate environmental interaction, especially in scenarios where direct force sensing is not practical. This paper focuses on deriving analytic formulations for the compliance of continuum robots that can be modeled as Kirchhoff rods. Compared to prior works, the approach presented herein is not subject to the constant-curvature assumptions to derive the configuration space compliance, and we do not rely on computationally-expensive finite difference approximations to obtain the task space compliance. Using modal approximations over curvature space and Lie group integration, we obtain closed-form expressions for the task and configuration space compliance matrices of continuum robots, thereby bridging the gap between constant-curvature analytic formulations of configuration space compliance and variable curvature task space compliance. We first present an analytic expression for the compliance of a single Kirchhoff rod. We then extend this formulation for computing both the task space and configuration space compliance of a tendon-actuated continuum robot. We then use our formulation to study the tradeoffs between computation cost and modeling accuracy as well as the loss in accuracy from neglecting the Jacobian derivative term in the compliance model. Finally, we experimentally validate the model on a tendon-actuated continuum segment, demonstrating the model's ability to predict passive deflections with error below 11.5\% percent of total arc length.
\end{abstract}
\section{Introduction} \label{sec:intro}
\par In this paper, we consider how to compute the passive compliance matrix of continuum 
 and soft robots modeled as Kirchhoff rods (i.e. Cosserat rods with negligible shear strains and extension). The \textit{local compliance} matrix provides a local prediction of the robot's passive deflections as a result of small changes in external loads, so it is useful for design and planning to ensure accurate physical interaction with the environment. The compliance matrix also needs to be computed for online passive stiffness modulation \cite{orekhov2019directional, rice2018passive,alamdari2018stiffness} and active stiffness/compliant motion control \cite{salisbury1980active,MahvashDupont2011Stiffness,goldman2014compliant}.
\par Many prior works presented mechanics models to predict the overall deflection of a continuum or soft robot for a given set of actuation and external loads \cite{burgner2015review,mahvash2011stiffness_control,oliverbutler2019prescribed}. However, relatively few studied the \emph{local} compliance, i.e. the small change in shape due to small changes in the applied forces \cite{gravagne2002ellipsoids,rucker2011computing,goldman2014compliant,black2017parallel}. These prior works on local compliance have defined the compliance matrix in two different ways: configuration space or task space compliance, depending on the application need. 
\par The \emph{configuration space compliance} relates external wrenches projected into the robot's configuration space to the ensuing changes in the configuration variables. This notion of compliance was used in \cite{goldman2014compliant} for compliant motion control, in \cite{Bajo2016hybrid,YasinSimaanIJRRSensing2021} for force regulation without dedicated end-effector force sensing, and in \cite{YasinSimaan2018SnakeScan} for force-controlled shape scanning of organs. Configuration space compliance can be computed analytically, but prior work has only presented it under the assumption of a constant-curvature shape.
\par The \emph{task space compliance} is the more conventional notion of compliance that provides the twist that a robot experiences as a result of a small change in an applied external wrench. While using finite differences to compute this compliance is possible, this approach is computationally expensive and does not provide analytic expressions of the compliance. Another method, applied to a concentric tube robot in \cite{rucker2011computing}, integrates an additional set of differential equations together with the standard Cosserat rod equations. This can be combined with finite difference steps to compute the compliance matrix of a parallel continuum robot\cite{black2017parallel}. In the mechanics literature, variational formulations were also used to derive stiffness matrices for geometrically exact mechanics models \cite{jelenic1995kinematically,simo1992symmetric}, but these methods have not been translated into a robotics context. Finally, \cite{gravagne2002ellipsoids} presented the task-space compliance of a continuum segment with discrete actuators applying moment loads along the segment's body. This work used modal basis functions to describe the backbone bending angle, resulting in an analytic expression for the task-space compliance, but it was limited to planar deflections. 
\par The contribution of this paper is a compliance matrix formulation that bridges the gap between constant-curvature configuration space compliance and geometrically-exact task space compliance. We present analytic expressions for both configuration and task space compliance. The analytic nature of the formulation enables sensitivity analysis of individual model parameters on the resulting configuration. As an example, we show how the term associated with the derivatives of the task space Jacobian and the external loading has a significant effect on the accuracy of the compliance model. We also show how our approach enables a tradeoff to be made to between model accuracy and computation cost, which is more difficult to achieve using prior geometrically exact mechanics models. 
\par In Section \ref{sec:kinematics}, we present the kinematic equations describing the variable curvature kinematic shape of the robot. In Section \ref{sec:single_rod}, we then show how to derive the task space compliance of a single Kirchhoff rod from the modal shape kinematics and local rod stiffness. We then extend this to the task space and configuration space compliance of a tendon-actuated continuum segment in Section \ref{sec:full_compliance_model}. Finally, in Section \ref{sec:results}, we present a simulation case study of a single Kirchhoff rod and an experimental validation of the compliance matrix on a tendon-actuated continuum segment.
\section{Lie Group Kinematics Preliminaries} \label{sec:kinematics}
\par In this section, we summarize the kinematic expressions that underpin our proposed compliance matrix formulation in this paper. The Lie group kinematics was presented in \cite{orekhov2023lie} as part of a formulation for continuum robot shape estimation from intrinsic string encoder measurements. Prior works have used modal shape functions to model hyper-redundant \cite{chirikjian1994modal} and continuum/soft robots \cite{jodicke2014lie,wang2014investigation,sadati2018ritz,gonthina2019euler,rao2021euler,sadati2022real}, but the kinematic expressions presented herein are most similar to those used in \cite{renda2020variable,boyer2020forward} for continuum robot mechanics, where the modal shape functions approximate the local backbone curvature.
\subsection{Central Backbone Kinematics}
\par
\begin{figure}[ht]
  \centering
  \includegraphics[width=\columnwidth]{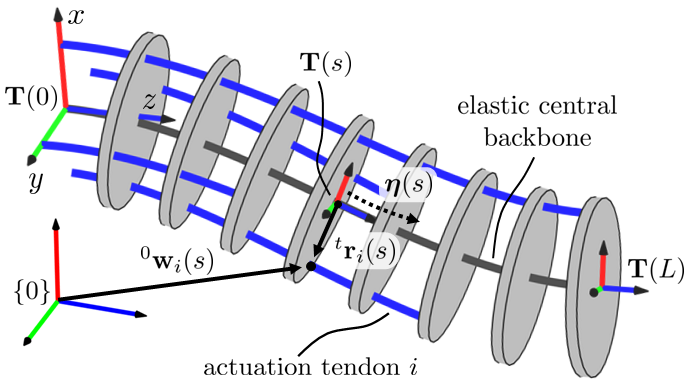}
  \caption{Kinematic parameters used in our modal-shape compliance matrix model.}
  \label{fig:kin_vars}
\end{figure}

 Referring to Fig.~\ref{fig:kin_vars}, we define the arc length distance along the central backbone as $s \in [0, L]$, where $L$ is the total length of the central backbone. We also assume the central backbone has a high slenderness ratio consistent with the assumption of negligible shear strains, i.e. a Kirchhoff rod. The central backbone can therefore be described by the curvature distribution along the backbone in three directions, \mbox{$\mb{u}(s) = [u_x,u_y,u_z]\T \in \realfield{3}$}. At a given arc length $s$, we assign a local body frame $\mb{T}(s) \in SE(3)$ with its z-axis tangent to the central backbone curve: 
\begin{equation} \label{eq:bb_frame_def}
\mb{T}(s) = \begin{bmatrix}
{}^0\mb{R}_t(s) & {}^0\mb{p}(s) \\
0 & 1
\end{bmatrix} \in SE(3), \quad s \in [0,L]
\end{equation}
where ${}^0\mb{R}_t(s)$ and ${}^0\mb{p}(s)$ are the orientation\footnote{We use the notation ${}^b\mb{R}_a$ to denote the orientation of frame \{A\} with respect to frame \{B\}. Also, frame \{A\} has its origin denoted by $\mb{a}$.} of the local body frame in the base frame \{0\} and the position of the origin of frame \{T\} in frame \{0\}. As the body frame traverses the central backbone curve, it undergoes a twist  \mbox{$\bs{\eta}(s) = [\mb{u}(s)\T,\mb{e}_3\T]\T \in \realfield{6}$}, where $\mb{e}_3 = [0,0,1]\T$ denotes the local tangent unit vector. Furthermore, it satisfies the following differential equation \cite{lynch2017modern}:
\begin{equation} \label{eq:bb_frame_deriv}
\begin{gathered}
 \mb{T}'(s) = \mb{T}(s)\widehat{\bs{\eta}}(s), \quad \widehat{\bs{\eta}}(s) = \begin{bmatrix}
 \widehat{\mb{u}}(s) & \mb{e}_3 \\
 0 & 0
 \end{bmatrix} \in se(3)
\end{gathered}
\end{equation}
where $\left( \cdot \right)'$ denotes the derivative with respect to $s$ and the hat operator $\left(\; \widehat{\cdot} \; \right)$ forms the standard matrix representations of $so(3)$ and $se(3)$ from the vector forms $\mb{u}$ and $\bs{\eta}$, respectively.
\par We now make a choice to express the curvature distribution $\mb{u}(s)$ as a weighted sum of polynomial functions. We denote the polynomial functions as $\bs{\phi}_x(s)$, $\bs{\phi}_y(s)$, and $\bs{\phi}_z(s)$ and the weights as $\mb{c}_x$, $\mb{c}_y$, and $\mb{c}_z$, for the $x$, $y$, and $z$ directions, respectively. The curvature $\mb{u}(s)$ is therefore:
\begin{equation} \label{eq:curvature_basis}
\mb{u}(s) = \begin{bmatrix} \bs{\phi}_x\T\mb{c}_x \\ \bs{\phi}_y\T\mb{c}_y \\ \bs{\phi}_z\T\mb{c}_z \end{bmatrix} =
\begin{bmatrix}
\bs{\phi}_x\T & 0 & 0 \\
0 & \bs{\phi}_y\T & 0 \\
0 & 0 & \bs{\phi}_z\T
\end{bmatrix} \begin{bmatrix} \mb{c}_x \\ \mb{c}_y \\ \mb{c}_z \end{bmatrix} = \bs{\Phi}(s)\mb{c}
\end{equation}
where the columns of $\bs{\Phi}(s) \in \realfield{3\times{}m}$ form a \emph{modal shape basis}, and $\mb{c} \in \realfield{m}$ is a vector of constant \emph{modal coefficients}.
\par We choose the Chebyshev polynomials of the first kind for the modal functions since their roots are also the Chebyshev nodes, resulting in optimal approximation. They can be computed recursively:
\begin{equation}\label{eq:cheby_recursive}
\begin{gathered}
T_0 = 1, \quad T_1(x) = x\\
T_n(x) = 2xT_{n-1}(x)-T_{n-2}(x), \quad n = 2,3,...
\end{gathered}
\end{equation}
where we shift the domain $x \in [-1, 1]$ to $s \in [0,L]$ via the transformation $x(s) = (2s - L)/L$. The benefit of this modal shape approach to modeling the kinematics of continuum robots is that variable curvature shapes with any desired choice of fidelity can be modeled by 
a suitable choice of the modal shape basis order. We will show below how this approach bridges the gap between prior constant-curvature configuration-space compliance and Kirchhoff rod task-space compliance models. 
\par For a given configuration $\mb{c}$, the body frame $\mb{T}(s)$ is found by integrating (\ref{eq:bb_frame_deriv}). As reviewed in \cite{iserles2000lie}, a variety of Lie group integration methods could be used for this, including an approach based on the Magnus expansion that we use here, following our result in~\cite{orekhov2020magnus}. After integrating (\ref{eq:bb_frame_deriv}), the spatial curve is given via a product of matrix exponentials:
\begin{equation} \label{eq:prod_of_exp}
\mb{T}(s) = \mb{T}(0)\prod_{i=0}^{k} e^{\mb{\Psi}_i}, \quad \mb{\Psi}_i \in se(3)
\end{equation}
where the particular form of $\mb{\Psi}_i$ depends on the choice of integration method \cite{iserles2000lie,orekhov2020magnus}. 
\subsection{Tendon Routing Kinematics}
To model the kinematics of the tendon paths, we follow the approach used in \cite{rucker2011general}. Assuming $p$ tendons, the tendon path is expressed in the moving frame $\mb{T}(s)$ and is given by:
\begin{equation} \label{eq:string_route_def}
{}^t\mb{r}_i(s) = [ r_{x_i}(s), r_{y_i}(s), 0] \T, \quad i = 1,2,\dots,p
\end{equation}
The position of a point along the tendon path is given in the world frame by:
\begin{equation} \label{eq:string_path}
{}^0\mb{w}_i(s) = {}^0\mb{p}(s) + \Rot{0}{t}(s)\,{}^t\mb{r}_i(s)
\end{equation}
Noting that vector norms are invariant under rotations, the length of the $i^{th}$ tendon is therefore given by:
\begin{equation} \label{eq:string_length}
\ell_i = \int_{0}^{s_{a_i}} \|{}^t\mb{w}'_i(s) \| \, ds, \quad i = 1,2,\dots,p \\
\end{equation}
where $s_{a_i}$ is the central backbone arc length at which the tendon is anchored to the spacer disk/end disk, and \mbox{${}^t\mb{w}'_i(s) = \Rot{0}{t}\T {}^0\mb{w}'_i(s)$} is found by taking the derivative of \eqref{eq:string_path} with respect to $s$ and substituting \eqref{eq:curvature_basis}, \mbox{${}^0\mb{p}'(s) = \Rot{0}{t}(s)\mb{e}_3$}, and ${}^0\mb{R}'_t(s) = \Rot{0}{t}(s)\widehat{\mb{u}}(s)$:
\begin{equation} \label{eq:string_path_deriv}
{}^t\mb{w}'_i(s)= \mb{e}_3 - {}^t\widehat{\mb{r}}_i(s)\mb{\Phi}(s)\mb{c}  + {}^t\mb{r}_i'(s), \quad i=1,2,\ldots p
\end{equation}
\subsection{Instantaneous Kinematics Jacobians}
\par Using the equations above, we also define two instantaneous kinematics Jacobians that will be used to formulate the configuration space and task space compliance matrices. Noting that the vector of modal coefficients $\mathbf{c}$ uniquely defines the configuration of the robot, we define the \emph{configuration space Jacobian} as the Jacobian relating small changes in the tendon lengths to small changes in the modal coefficients:
\begin{equation} \label{eq:config_jacobian}
d\bs{\ell} = \mb{J}_{\ell{}c}d\mb{c}, \quad \mb{J}_{\ell{}c} \in \realfield{p \times m}
\end{equation}
\par\noindent $\mb{J}_{\ell{}c}$ is found by taking derivatives of \eqref{eq:string_length}. We also define the \emph{body Jacobian} relating small changes in the modal coefficients to body frame changes at arc length $s$:
\begin{equation} \label{eq:body_jacobian}
\bs{\xi}(s) = \mb{J}_{\xi{}c}(s)d\mb{c}, \quad \bs{\xi} \in se(3)
\end{equation}
$\mb{J}_{\xi{}c}(s)$ is found via the derivative of the exponential mapping together with \eqref{eq:prod_of_exp}. 
\section{Compliance Matrix of a Kirchhoff Rod} \label{sec:single_rod}
\par We now build on the kinematic equations above to formulate the compliance matrix of a Kirchhoff rod. We follow a similar set of steps as the derivation of the stiffness matrix for a parallel robot in \cite{orekhov2019directional} and the statics \cite{simaan2005snake} and stiffness of multi-backbone robots \cite{goldman2014compliant}. We begin by defining a small perturbation (i.e. a twist) in the end effector pose $\mb{T}(L)$, denoted as $\delta\mb{x}_h$, which we find by computing the body twist and then transforming into the hybrid frame that is coincident with the body frame $\mb{T}(L)$ but aligned with the world frame:
\begin{equation}
\begin{gathered}
\delta\mb{x}_h = {}^h\mb{S}_b\left(\mb{T}^{-1}(L)\delta\mb{T}(L)\right)^\vee \in \realfield{6} \\
{}^h\mb{S}_b = \begin{bmatrix}
\Rot{w}{b} & \mb{0}_{3x3} \\
\mb{0}_{3x3} & \Rot{w}{b}
\end{bmatrix} \in \realfield{6 \times 6}
\end{gathered}
\end{equation}
where $\Rot{w}{b} \in SO(3)$ is the rotation matrix of the body frame expressed in the world frame and the ``vee" operator $\left(^\vee\right)$ extracts a vector $\mb{x}$ from its corresponding skew-symmetric cross-product matrix $\mb{x}^\wedge\in se(3)$. We then define the task-space compliance matrix of a single rod as:
\begin{equation} \label{eq:body_compliance_def}
\delta\mb{x}_h = \mb{C}_{x_{rod}}\delta{}\mb{w}_{h}, \quad \mb{C}_{x_{rod}} \in \realfield{6 \times 6}
\end{equation}
where $\delta\mb{w}_{h} = [\delta\mb{m}_h\T,\delta\mb{f}_h\T]\T \in \realfield{6}$ is a small change in the applied wrench written in the hybrid frame\footnote{The hybrid frame has its origin coincident with the origin of the local frame \{T\} and its axes parallel to the base frame \{0\}.} and with the moment followed by the force.
\par We assume here the rod is massless, and that shear strains and extension are negligible. We denote $\mb{K}_{bt} = \text{diag}(EI_x,EI_y,JG) \in \realfield{3\times 3}$ as the diagonal bending/torsion stiffness matrix of the rod's cross section. The bending energy in the rod is given by
\begin{equation} \label{eq:bending_energy}
E = \int_0^L \frac{1}{2} \mb{u}(s)\T\mb{K}_{bt}\mb{u}(s) \,\text{d}s = \frac{1}{2}\mb{c}\T \underbrace{\left( \int_0^L \bs{\Phi}\T\mb{K}_{bt}\bs{\Phi} \, \text{d}s \right)}_{\bs{\Phi}_k} \mb{c}
\end{equation}
where we have substituted the modal shape basis curvature from \eqref{eq:curvature_basis}. Note that the matrix $\bs{\Phi}_k$ can be computed offline if the modal shape functions are chosen \emph{a priori}. 
\par For the work done by the applied wrench as it produces a small displacement $\delta\mb{x}$, there is a corresponding small change in bending energy $\delta{}E$:
\begin{equation} \label{eq:virtual_rod_work}
\mb{w}_h\T\delta\mb{x}_h = \delta{}E
\end{equation}
Recalling the body Jacobian from \eqref{eq:body_jacobian}, we denote the relationship between $\delta\mb{x}_h$ and $\delta\mb{c}$ as:
\begin{equation}\label{eq:delta_xh}
\delta\mb{x}_h = {}^h\mb{S}_b\mb{J}_{\xi c} \delta\mb{c} = \widetilde{\mb{J}}_{\xi c} \delta\mb{c}
\end{equation}
and substitute this into \eqref{eq:virtual_rod_work}:
\begin{equation}
\mb{w}_h\T\widetilde{\mb{J}}_{\xi c}\delta\mb{c} = \left( \frac{\partial{}E}{\partial\mb{c}} \right)\T\delta\mb{c}
\end{equation}
By the principle of virtual work, to be in static equilibrium we require the virtual displacements associated with $\delta\mb{c}$ to vanish, resulting in:
\begin{equation}
\widetilde{\mb{J}}_{\xi{} c}\T\mb{w}_h = \frac{\partial{}E}{\partial\mb{c}}
\end{equation}
Denoting $c_i$ as the $i^{th}$ element of $\mb{c}$ and taking small perturbations about the equilibrium configuration:
\begin{equation}
\delta \left( \widetilde{\mb{J}}_{\xi{}c}\T \right) \mb{w}_h + \widetilde{\mb{J}}_{\xi{}c}\T \delta\mb{w}_h = \delta \left( \frac{\partial{}E}{\partial\mb{c}} \right)
\end{equation}
By substituting $\delta \left(\widetilde{\mb{J}}_{\xi{}c}\T\right) =\displaystyle\sum_{i=1}^n \left[\frac{\partial\widetilde{\mb{J}}_{\xi{}c}\T}{\partial{}c_i}\right] \delta c_i$ and $\delta \left( \frac{\partial{}E}{\partial\mb{c}} \right) =\left[\frac{\partial{}^2E}{\partial\mb{c}^2}\right]\delta\mb{c}$ and solving for $\delta \mb{c}$ we obtain:
\begin{equation} \label{eq:dc_single_rod}
\delta\mb{c}=\left( \frac{\partial{}^2E}{\partial\mb{c}^2} - \left[ \frac{\partial\widetilde{\mb{J}}_{\xi{}c}\T}{\partial{}c_1}\mb{w}_h \dots \frac{\partial\widetilde{\mb{J}}_{\xi{}c}\T}{\partial{}c_{n}}\mb{w}_h \right] \right)^{-1} \widetilde{\mb{J}}_{\xi{}c}\T\delta\mb{w}_h
\end{equation}
Recalling \eqref{eq:body_compliance_def} and substituting \eqref{eq:dc_single_rod} into $\delta\mb{x}_h = \widetilde{\mb{J}}_{\xi{}c}\delta\mb{c}$ results in the analytic expression for the compliance matrix:
\begin{equation}\label{eq:single_rod_Cx}
\mb{C}_{x_{rod}} = \widetilde{\mb{J}}_{\xi{}c}\left( \frac{\partial{}^2E}{\partial\mb{c}^2} - \left[ \frac{\partial\widetilde{\mb{J}}_{\xi{}c}\T}{\partial{}c_1}\mb{w}_h \dots \frac{\partial\widetilde{\mb{J}}_{\xi{}c}\T}{\partial{}c_{n}}\mb{w}_h \right] \right)^{-1} \widetilde{\mb{J}}_{\xi{}c}\T
\end{equation}
The above result matches with the congruence transformation of stiffness as discussed in \cite{chen2000conservative} for serial robots. The energy Hessian $\frac{\partial{}^2E}{\partial{}\mb{c}^2} $, which can be computed offline, is found by differentiating (\ref{eq:bending_energy}) twice:
\begin{equation}
\frac{\partial{}E}{\partial{}\mb{c}} = \frac{1}{2}\left( \bs{\Phi}_k + \bs{\Phi}_k\T  \right) \mb{c}~~ \Rightarrow ~~
\frac{\partial{}^2E}{\partial{}\mb{c}^2} = \frac{1}{2}\left( \bs{\Phi}_k + \bs{\Phi}_k\T  \right)
\end{equation}
%

\section{Compliance Matrices of a Tendon-actuated Continuum Segment} \label{sec:full_compliance_model}
\par We now extend the example above for a single rod to the case of a tendon-actuated continuum segment. We first derive a statics model similar to the constant curvature model in \cite{simaan2004dexterous} but for a variable curvature segment. We then use this statics model to arrive at the task-space compliance matrix, in an analytic form as a result of our Lie group modal shape formulation. Finally, we present the variable curvature configuration-space compliance matrix.
\subsection{Task-space compliance}
\par For a given external wrench $\mb{w}_h$ on the tip of the segment and a perturbation $\delta\mb{x}_h$ in the end pose, we have a corresponding change in the forces applied to the actuation tendons $\bs{\tau}$ and a corresponding change in the bending energy stored in the segment, where the bending energy is given by \eqref{eq:bending_energy}. Following the principle of virtual work, we have:
\begin{equation}
\mb{w}_h\T\delta\mb{x}_h + \bs{\tau}\T\delta\bs{\ell} - \delta E = 0
\end{equation}
where $\delta\bs{\ell}$ is a change in the tendon length. Referring to \eqref{eq:body_jacobian} and \eqref{eq:config_jacobian}, we substitute \eqref{eq:delta_xh} and \mbox{$\delta E = \frac{\partial{}E}{\partial\mb{c}}\T\delta\mb{c}$} to arrive at the statics of the segment about a given configuration:
\begin{equation} \label{eq:var_curv_statics}
\mb{J}_{\ell c}\T \bs{\tau} = \frac{\partial{}E}{\partial\mb{c}} - \widetilde{\mb{J}}_{\xi c}\T\mb{w}_h
\end{equation}
We then take small perturbations of this statics expression:
\begin{equation} \label{eq:compliance_step3}
\mb{C}_\tau\delta\mb{c} + \mb{J}_{\ell c}\T \delta\bs{\tau} = \frac{\partial^2 E}{\partial\mb{c}^2}\delta\mb{c} - \mb{C}_{w_h} \delta\mb{c} - \widetilde{\mb{J}}_{\xi c}\T \delta\mb{w}_h
\end{equation}
where $\mb{C}_\tau$ and $\mb{C}_{w_h}$ are defined as:
\begin{equation}
\mb{C}_\tau = \left[\frac{\partial\mb{J}_{\ell c}\T}{\partial c_1} \bs{\tau},\ldots,\frac{\partial\mb{J}_{\ell c}\T}{\partial c_n} \bs{\tau}\right] \in \realfield{n \times n}
\end{equation}
\begin{equation}
\mb{C}_{w_h} = \left[ \frac{\partial\widetilde{\mb{J}}_{\xi c}\T}{\partial c_1} \mb{w}_h,\ldots,\frac{\partial\widetilde{\mb{J}}_{\xi c}\T}{\partial c_n} \mb{w}_h \right]  \in \realfield{n \times n}
\end{equation}
The matrix $\mb{C}_\tau$ is the contribution to the compliance matrix of the forces on the tendons at the current configuration, and $\mb{C}_{w_h}$ is the contribution due to the external wrench.   $\mb{C}_\tau$ can be readily affected by using actuation redundancy (internal preload) and $\mb{C}_{w_h}$ depends only on the external load. In a robot without actuation redundancy, $\bs{\tau}$ is determined by $\mb{w}_h$ through the statics equation. Therefore $\mb{C}_\tau$ and $\mb{C}_{w_h}$ are not completely independent.  In a robot with actuation redundancy, these two matrices are independent. 
\par By defining the \textit{joint-level stiffness} $\mb{K}_\ell$ as a diagonal stiffness matrix containing the stiffness of individual actuation lines, i.e., \mbox{$\mathbf{K}_\ell(i,i) = \frac{\delta \tau_{i}}{\delta \ell_i}$}, we can use $\delta\bs{\tau} = \mb{K}_{\ell}\delta\bs{\ell}$ and $\delta\bs{\ell} = \mb{J}_{\ell c}\delta\mb{c}$ in \eqref{eq:compliance_step3}, combine like terms, and solve for $\delta\mb{c}$:
\begin{equation}
 \delta \mb{c} = \left(\frac{\partial^2 E}{\partial\mb{c}^2} - \mb{C}_\tau - \mb{C}_{w_h} - \mb{J}_{\ell c}\T\mb{K}_\ell\mb{J}_{\ell c} \right)^{-1}\widetilde{\mb{J}}_{\xi c}\T\delta\mb{w}_h
\end{equation}
We then substitute this expression into $\delta\mb{x}_h = \widetilde{\mb{J}}_{\xi c}\delta\mb{c}$ to arrive at the compliance matrix:
\begin{equation} \label{eq:tendon_task_space_compliance}
\mb{C}_x = \frac{\delta\mb{x}_h}{\delta \mb{w}_h} = \widetilde{\mb{J}}_{\xi c}\left(\frac{\partial^2 E}{\partial\mb{c}^2} - \mb{C}_\tau - \mb{C}_{w_h} - \mb{J}_{\ell c}\T\mb{K}_\ell\mb{J}_{\ell c} \right)^{-1}\widetilde{\mb{J}}_{\xi c}\T
\end{equation}
As a result of our analytic kinematic expressions, \eqref{eq:tendon_task_space_compliance} bears resemblance to the results of prior works on stiffness modulation of rigid-link parallel robots, where one defines an active stiffness term dependent on the joint-level forces and the derivative of the Jacobian. and a passive stiffness term dependent on joint-level stiffness \cite{yi_open-loop_1989,yi1993geometric,simaan2003geometric}. Compared to a rigid-link parallel robot stiffness matrix model, additional terms for tendon-actuated continuum robots include the Hessian of the bending energy and a term with the joint forces multiplied by the derivatives of the configuration-space Jacobian.
\subsection{Configuration-space compliance}
\par We now present the configuration space compliance matrix for a tendon-actuated continuum segment with variable curvature deflections. Our formulation can be seen as an extension of the compliance matrix in \cite{goldman2014compliant} to the case of variable curvature continuum robots. The resulting expression does not require computation of the task space Jacobian and its derivatives, and is therefore less computationally expensive. Furthermore, it does not require knowledge of the external wrench as required by the task-space compliance, so we believe it can be used for compliant motion control as done in \cite{goldman2014compliant}, but for cases where large external loads produce variable curvature deflections.
\par Referring to \eqref{eq:var_curv_statics}, we first define the external wrench projected into the configuration space, denoted as $\mb{w}_c$:
\begin{equation} \label{eq:c_space_wrench}
\mb{w}_c = \widetilde{\mb{J}}_{\xi c}\T\mb{w}_h = \frac{\partial{}E}{\partial\mb{c}} - \mb{J}_{\ell c}\T \bs{\tau}
\end{equation}
Taking small perturbations of \eqref{eq:c_space_wrench} results in
\begin{equation}
\delta\mb{w}_c = \frac{\partial^2{}E}{\partial\mb{c}^2}\delta\mb{c} - \mb{C}_\tau\delta\mb{c} - \mb{J}_{\ell c}\T \delta\bs{\tau}
\end{equation}
where $\mb{C}_\tau$ was given in \eqref{eq:compliance_step3}. We now substitute $\delta\bs{\tau} = \mb{K}_{\ell}\mb{J}_{\ell c}\delta\mb{c}$, combine like terms, and solve for $\delta\mb{c}$:
\begin{equation}
\delta\mb{c} = \left( \frac{\partial^2{}E}{\partial\mb{c}^2} - \mb{C}_\tau - \mb{J}_{\ell c}\T \mb{K}_{\ell}\mb{J}_{\ell c} \right)^{-1}\delta\mb{w}_c
\end{equation}
Computing the change in the configuration for a small change in the projected wrench, i.e. $\delta \mb{c} = \mb{C}_c\delta \mb{w}_c$, results in the following configuration-space compliance matrix:
\begin{equation} \label{eq:config_compliance}
\mb{C}_c = \left( \frac{\partial^2{}E}{\partial\mb{c}^2} - \mb{C}_\tau - \mb{J}_{\ell c}\T \mb{K}_{\ell}\mb{J}_{\ell c} \right)^{-1}, \;\; \mb{C}_c \in \realfield{n \times n}
\end{equation}
Note that \eqref{eq:config_compliance} does not require that the task-space Jacobian and its derivatives be computed. It also does not require knowledge of the external wrench. It does, however, in the general case require knowledge of the actuation tendon forces. These forces can come either directly from force sensors placed in series with the actuation tendons \cite{goldman2014compliant,black2017parallel} or from estimates obtained via the commanded motor current. In special cases where $\mb{J}_{\ell c}$ constant, e.g. segments with constant pitch tendon routing and negligible torsional deflections as in Section \ref{sec:results}, $\mb{C}_\tau$ is zero and the actuation tendon forces do not need to be measured.
\subsection{Determining the configuration modal coefficients}
\par Both the configuration-space and task-space compliance matrices require the modal coefficients $\mb{c}$ to define the segment's configuration. There are two ways to obtain $\mb{c}$. The first is using the shape sensing approach presented in \cite{orekhov2023lie}. This approach requires integration of shape-sensing hardware, but the benefit is that it does not require a potentially computationally expensive mechanics model. 
\par The second way to obtain $\mb{c}$ is with a mechanics model like the ones presented in \cite{orekhov2020magnus,oliverbutler2019prescribed}. After solving the mechanics model, $\mb{c}$ can be obtained by converting the collocation values into modal coefficients, as shown in \cite{orekhov2017modeling}. Using a mechanics model to obtain $\mb{c}$ does not require shape-sensing, but does require an estimate of the external wrench applied to the segment, which may be known \emph{a-priori} in some cases or can be measured using a load cell attached to the end effector.
\section{Simulation and Experimental Results} \label{sec:results}
In this section, we present a simulation-based analysis of the Kirchhoff rod compliance matrix model, as well as an experimental validation and analysis of the compliance matrix for a tendon-actuated continuum segment. 
\subsection{Kirchhoff rod model validation and analysis in simulation}
\par We first compare the deflections predicted by our compliance matrix in \eqref{eq:single_rod_Cx} to the deflections predicted by a Kirchhoff rod model, solved following the method in \cite{orekhov2017modeling} which combined orthogonal collocation and Lie group integration. We considered a 2\,mm diameter Nitinol rod with a length of 200\,mm and combinations of $\pm$1\,N tip forces and $\pm$0.5\,Nm tip wrenches, generating a set of 2187 rod shapes. For each shape, we applied small steps of 0.1\,N and 0.05\,Nm to increment the applied force/moment (6 deflections for each shape) and recomputed the Kirchhoff rod model. We then used \eqref{eq:single_rod_Cx} to predict the deflection for each wrench increment. For \eqref{eq:single_rod_Cx}, we used the same Chebyshev series for the $x$, $y$, and $z$ directions, but varied the order of the Chebyshev series from $n=0$ to $n=10$, i.e. when $n=0$, $\mb{\Phi}$ had three columns, and when $n=10$, $\mb{\Phi}$ had 33 columns. We used 10 integration steps when computing $\mb{T}(L)$ regardless of $n$. We then computed the tip translational deflection error:
\begin{equation} \label{eq:pos_err}
e_p = \| \Delta \mb{p}_{gt} - \Delta\mb{p}_m \| \in \realfield{}
\end{equation}
where $\Delta\mb{p}_{gt} \in \realfield{3}$ is the deflection predicted by the Kirchhoff rod model and $\Delta\mb{p}_{m} \in \realfield{3}$ is the deflection from \eqref{eq:single_rod_Cx}. The solvers were run using MATLAB 2021a on an Intel i7-11700 CPU.
\begin{figure}[ht]
  \centering
  \includegraphics[width=\columnwidth]{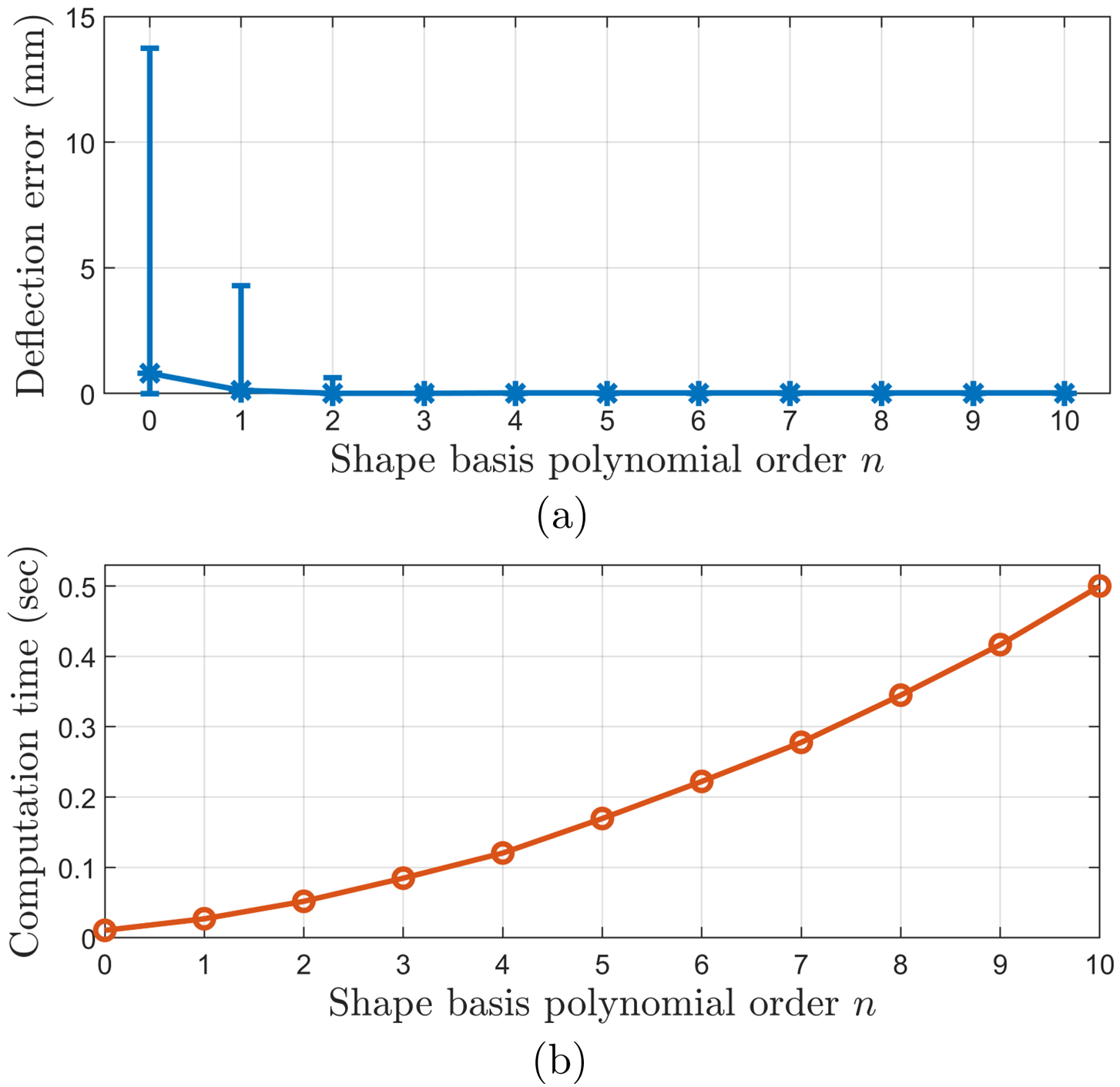}
  \caption{(a) The absolute deflection error in tip deflection, with error bars showing the minimum and maximum error, predicted by our compliance matrix expression and (b) the computation time for different numbers of modal coefficients, showing a trade-off between speed and accuracy.}
  \label{fig:single_rod_comp_err}
\end{figure}
\begin{table}[ht]\caption{Tip Deflection Error versus Polynomial Order} \label{table:single_rod_comp_errors}
\centering \begin{tabular} {|c|C{0.95cm}|C{0.95cm}|C{0.95cm}|C{0.95cm}|C{1.3cm}|} \hline
         & \multicolumn{2}{c|}{Position error (mm)} & \multicolumn{2}{c|}{Rotation error (deg)} & \\ \hline
         & Avg.            & Max.               & Avg.           & Max.          & Speed (Hz)  \\ \hline
$n = 0$  &  0.8           & 13.7              & 0.2           & 6.3          & 94.6        \\  \hline
$n = 2$  &  1.3\text{e-2}  & 0.6               & 2.2\text{e-3}  & 0.10          & 19.3        \\ \hline
$n = 4$  &  6.7\text{e-5}  & 6.3\text{e-3}      &  2.0\text{e-5} & 1.2\text{e-3} & 8.3         \\ \hline
$n = 6$  &  1.5\text{e-5}  & 3.8\text{e-4}      & 1.3\text{e-5}  & 4.4\text{e-4} & 4.5         \\ \hline
\end{tabular}
\end{table}
\par The mean/max absolute deflection error results are shown in Table \ref{table:single_rod_comp_errors} and Fig. \ref{fig:single_rod_comp_err}. We see that as $n$ is increased, the analytic expression rapidly converges to the simulated Kirchhoff rod deflection. We also observe that computation time increases with $n$, showing a tradeoff between the compliance matrix accuracy and computation cost.
\par The primary source of computational cost is in computing the derivatives of $\mb{J}_{\xi c}$. For $n=10$, estimating these derivatives accounts for approximately 95\% of the computation time. We are currently using finite differences to estimate these derivatives, but we believe it is possible to derive these derivatives analytically and reduce computation cost. Methods from \cite{iserles2000lie} to reduce the number of Lie brackets needed to calculate $\mb{J}_{\xi c}$ could also reduce computation cost.
\par A benefit of our formulation, in contrast to other approaches to computing the compliance matrix, is that it allows a trade-off between computation cost and accuracy to be made depending on the application need. A high-order model (large $n$) to be used when computing the statics model to accurately predict the spatial shape of the continuum robot, but use a lower-order model (small $n$) to compute the compliance matrix for online compliant motion control or predicting local deflections.
\begin{table}[ht]\caption{Tip Deflection Error versus Polynomial Order When Neglecting the Jacobian Derivative Term} \label{table:no_jac_deriv}
\centering \begin{tabular} {|c|C{0.95cm}|C{0.95cm}|C{0.95cm}|C{0.95cm}|C{1.3cm}|} \hline
         & \multicolumn{2}{c|}{Position error (mm)} & \multicolumn{2}{c|}{Rotation error (deg)} & \\ \hline
         & Avg.   & Max. & Avg. & Max.   & Speed (Hz)  \\ \hline
$n = 0$  &  3.23  & 30.5 & 0.94 &  14.3  & 665.3        \\  \hline
$n = 2$  &  2.82  & 37.1 & 1.05 &  10.1  & 335.0        \\ \hline
$n = 4$  &  2.83  & 37.1 & 1.06 &  10.2  & 237.0        \\ \hline
$n = 6$  &  2.83  & 37.1 & 1.06 &  10.2  & 184.8         \\ \hline
\end{tabular}
\end{table}
\par To reduce computation cost, one may ask whether it is possible to neglect the term that includes the derivatives of the task-space Jacobian and external wrench. While this term is typically negligible for stiff rigid-link robots \cite{alici2005enhanced}, it was shown in \cite{pitt_investigation_2015} that this term is not negligible for robots with compliant actuators. Here we show a similar result for continuum robots, by computing \eqref{eq:single_rod_Cx} while neglecting this term using the same set of 2187 rod shapes and increments in wrench described above, and comparing again to the local deflections predicted by the Kirchhoff rod model. The mean absolute position and orientation errors are given in Table. \ref{table:no_jac_deriv}. We observe that while the speed of computing \eqref{eq:single_rod_Cx} increases significantly, the maximum position and orientation errors are high when neglecting this term. We also observe that increasing the number of modal coefficients does not significantly improve the modeling error when neglecting the task space Jacobian derivatives.
\subsection{Experimental validation}
\par
\begin{figure}[ht]
\centering
  \includegraphics[width=\columnwidth]{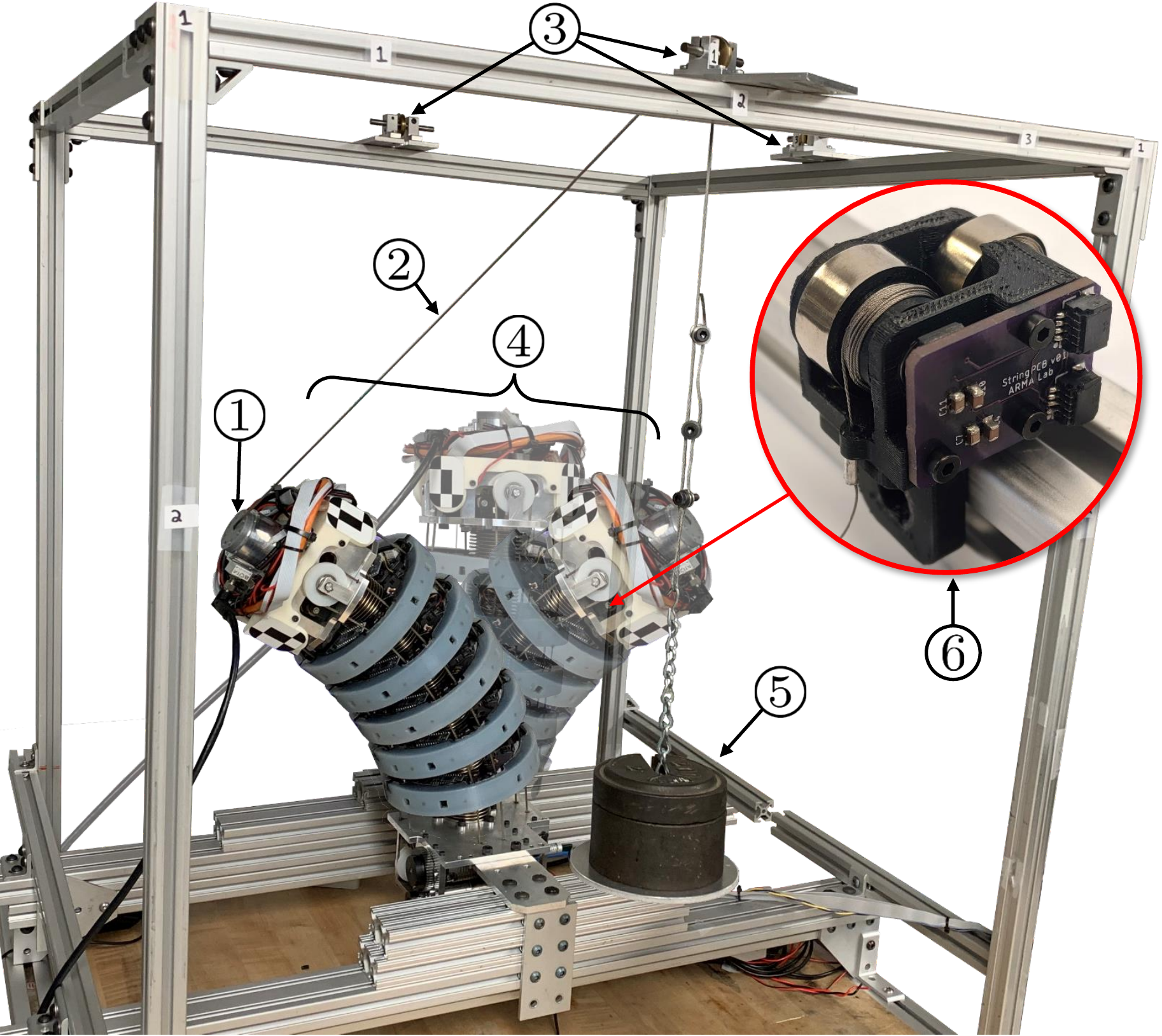}
  \caption{The setup used to experimentally validate the compliance model: \protect\circled{1} force/torque sensor, \protect\circled{2} wire-rope, \protect\circled{3} pulleys, \protect\circled{4} three out of five configurations used in the experiments, and \protect\circled{5} mass used to apply forces to the end disk. The end disk of the robot has four string encoders \protect\circled{6} that measure the segment's configuration.}
\label{fig:experimental_setup}
\end{figure}

 In this section, we describe the experimental validation of our method on a tendon-actuated continuum segment, shown in Fig. \ref{fig:experimental_setup}. The continuum segment central backbone is 300.6\,mm in length and the kinematic radius of the actuation tendons is 65.2\,mm. Two actuators are integrated into the base, which drive the actuation tendons that bend the segment. In the end-disk are four string encoders that measure the deflections of the segment to estimate $\mb{c}$. In \cite{orekhov2023lie}, it was shown that this shape sensing approach, which is purely kinematic and does not require a mechanics model, can estimate the end disk position to within 5\% of total arc length, or 14.4\,mm of position error.
\par Metal bellows in the segment's flexible backbone structure make it approximately 1950 times stiffer in torsion that in bending, so we neglect the torsional deflections and use the following modal shape basis:
\begin{equation}
\mb{u}(s) = \begin{bmatrix} 
\bs{\phi}_x\T & 0 \\ 
0 & \bs{\phi}_y\T \\ 
0 & 0 
\end{bmatrix}
\renewcommand*{\arraystretch}{1.1}
\begin{bmatrix}
\mb{c}_x \\ \mb{c}_y
    \end{bmatrix} = \mb{\Phi}(s)\mb{c}, \quad \mb{c} \in \realfield{6}
\end{equation}
where $\bs{\phi}_x\T(s) = \bs{\phi}_y\T = \begin{bmatrix} T_0 & T_1 & T_2 \end{bmatrix}\T$. With this choice of $\mb{\Phi}$ and since the strings and actuaton tendons are routed in constant-pitch radius paths, $\mb{J}_{\ell c}$ is constant, and $\mb{C}_\tau = \mb{0}$.
\begin{table}[ht]\caption{Joint-space  configurations used in experiments} \label{table:joint_space_configs}
\centering \begin{tabular} {|c|c|c|c|c|c|}  \hline
 Configuration & 1 & 2 & 3 & 4 & 5 \\ \hline 
$\theta_1$  &  0\textdegree  & 500\textdegree & -500\textdegree & 500\textdegree & 0\textdegree  \\  \hline
$\theta_2$  & 500\textdegree  & 0\textdegree & 500\textdegree  & -500\textdegree & 0\textdegree        \\ \hline
\end{tabular}
\end{table}
\par The experimental setup used to validate the compliance matrix model is shown in Fig. \ref{fig:experimental_setup}. In order to experimentally validate our approach, we attached a force/torque sensor (Bota Systems Rokubi\texttrademark) to the end effector of the segment. The segment was actuated to five different joint space configurations in the segment's workspace. The angles of the first and second actuators, denoted as $\theta_1$ and $\theta_2$, respectively, are given in Table \ref{table:joint_space_configs}. At each configuration, the unloaded configuration of the segment was measured using the string encoders. A mass was then attached to the force/torque sensor via a wire-rope hung over a pulley. The segment's new deflected configuration and the applied wrench was then measured. This was repeated for each configuration using 10 different combinations of wire rope direction and masses ranging from 2\,lbs to 10\,lbs, resulting in a total of 50 measured deflections. 
\par Across the 50 deflections, the mean and maximum deflection was 14.3\,mm and 66.7\,mm, respectively. The first 10 deflections used for the first joint-space configuration were used to calibrate the model parameters $EI_x = EI_y$ and $k_1 = k_2$. We did this by solving for the values of these parameters that minimized the least-squared error between the model predicted deflection and the experimentally measured deflection, solved with \emph{lsqnonlin()} in MATLAB. The resulting calibrated values were $EI_x = EI_y = 3.20$\,Nm$^2$ and $k_1 = k_2 = 1.23\text{e}5$\,N/m.
 \begin{figure}[ht]
  \centering
  \includegraphics[width=0.99\columnwidth]{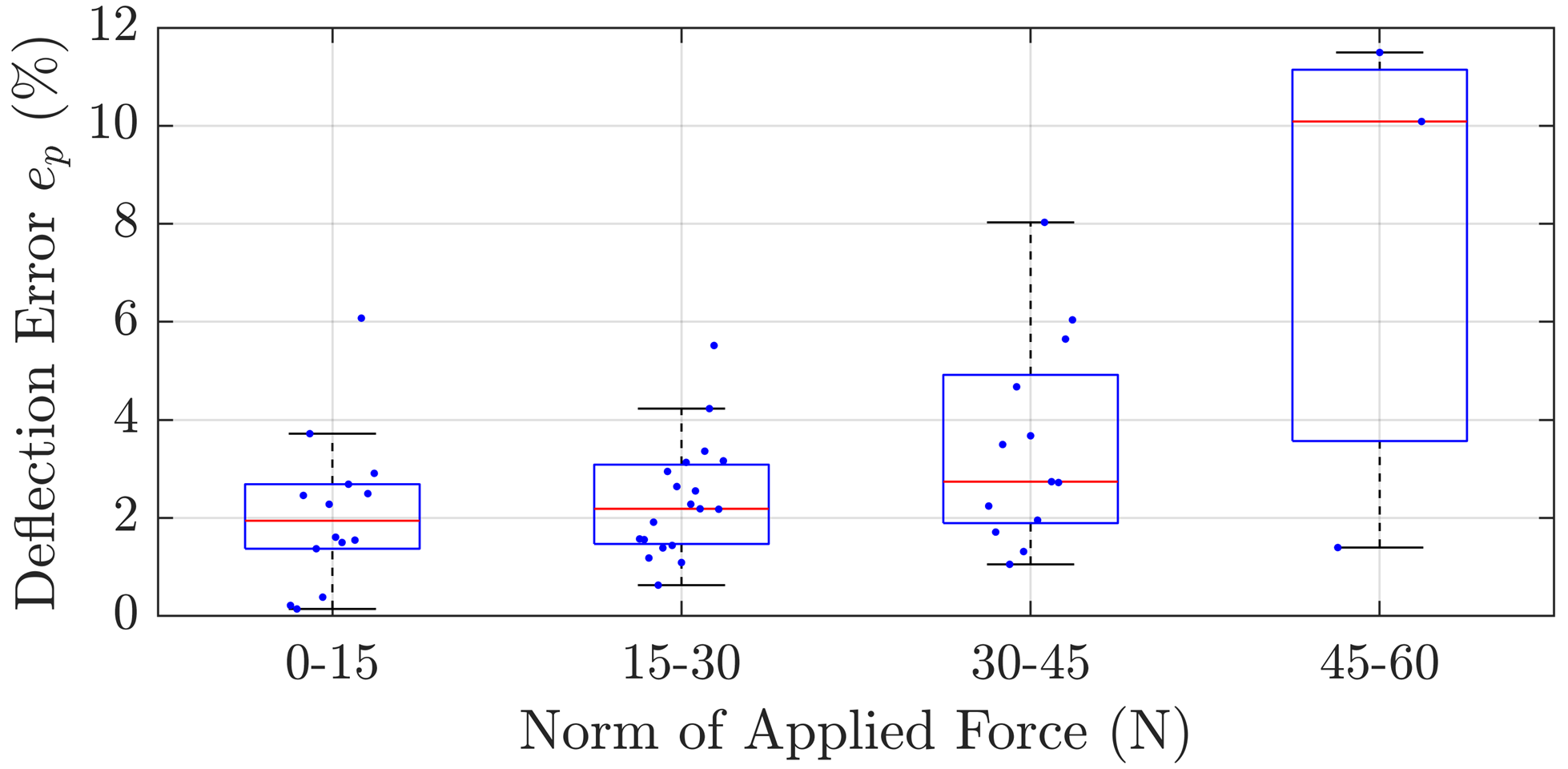}
  \caption{Boxplot comparing deflection error to the norm of the force applied to the end disk. The central red line is the median, and the box covers the 25th and 75th percentiles. Small dots represent the experimental data.}
  \label{fig:results}
\end{figure}
\par The remaining 40 deflections were used to validate the model. Comparing the measured and model-predicted deflections, the mean and max absolute error $e_p$ from \ref{eq:pos_err} was 9.1\,mm and 34.6\,mm, respectively, or 3.0\% and 11.5\% of the backbone length. Since the compliance matrix is only a local estimate of the deflection behavior, it is expected that larger deflections result in larger error in the predicted deflection. Figure \ref{fig:results} shows predicted deflection error for different norms of applied force. We observe that, as expected, predicted deflection error tends to increase as the applied force increases.
\section{Conclusions}
\par In this paper, we presented a method for computing the compliance matrix of continuum and soft robots utilizing a modal shape basis Lie group formulation. Compared to prior work, our approach does not rely on a constant-curvature assumption and leads to analytic expressions for both configuration space and task space compliance. We presented the compliance of a single Kirchhoff rod, and we highlighted through a simulation study the tradeoff between computational cost and modeling accuracy as well as the importance of including the Jacobian derivatives when computing compliance. We also presented the compliance for a tendon-actuated continuum segment, and performed an experimental validation of the task-space compliance, showing predicted deflection errors below 11.5\% of arc length. Future work includes applying this formulation to passive stiffness modulation and compliant motion control of variable curvature continuum and soft robots.
\bibliographystyle{IEEEtran}
\bibliography{bib/thesis_refs,bib/additional_refs}
\end{document}